\documentclass[a4paper, 10pt, conference]{ieeeconf}      

\IEEEoverridecommandlockouts                              

\usepackage{graphics} 
\usepackage{epsfig} 
\usepackage{mathptmx} 
\usepackage{times} 
\usepackage{amsmath} 
\usepackage{amssymb}  
\usepackage{hyperref}
\usepackage{hyphenat}
\usepackage{color, colortbl}
\usepackage{multirow}
\usepackage{gensymb}

\title{\LARGE \bf
NeurAll: Towards a Unified Visual Perception Model \\ for Automated Driving   
}

\author{Ganesh Sistu$^{1}$, Isabelle Leang$^{2}$, Sumanth Chennupati$^{3,5}$, Senthil Yogamani$^{1}$, \\ Ciar\'{a}n Hughes$^{1}$, Stefan Milz$^{4}$ and Samir Rawashdeh$^{5}$ \\ 
$^{1}$Valeo Vision Systems, Ireland $^{2}$Valeo Bobigny, France $^{3}$Valeo Troy, United States \\ $^{4}$Valeo Kronach, Germany  $^{5}$University of Michigan-Dearborn, United States\\
         {\tt \small \{ganesh.sistu,isabelle.leang,senthil.yogamani,ciaran.hughes,stefan.milz\}@valeo.com } \\
         {\tt \small \{schenn,srawa\}@umich.edu}
}

\begin{document}

\maketitle
\thispagestyle{empty}
\pagestyle{empty}

\begin{abstract}

Convolutional Neural Networks (CNNs) are successfully used for the important automotive visual perception tasks including object recognition, motion and depth estimation, visual SLAM, etc.  However, these tasks are typically independently explored and modeled. In this paper, we propose a joint multi-task network design for learning several tasks simultaneously. 
Our main motivation is the computational efficiency achieved by sharing the expensive initial convolutional layers between all tasks. Indeed, the main bottleneck in automated driving systems is the limited processing power available on deployment hardware.
There is also some evidence for other benefits in improving accuracy for some tasks and easing development effort. 
It also offers scalability to add more tasks leveraging existing features and achieving better generalization. We survey various CNN based solutions for visual perception tasks in automated driving. Then we propose a unified CNN model for the important tasks and discuss several advanced optimization and architecture design techniques to improve the baseline model.
The paper is partly review and partly positional with demonstration of several preliminary results promising for future research. 
We first demonstrate results of multi-stream learning and auxiliary learning which are important ingredients to scale to a large multi-task model. Finally, we implement a two-stream three-task network which performs better in many cases compared to their corresponding single-task models, while maintaining network size.

\end{abstract}

\section{Introduction} \label{intro}

Automated driving is a rapidly advancing application domain, and is being made feasible due to advances in various technology areas, including perception, mapping, fusion, contextualisation, path planning and control. On the perception side, cameras are a dominant sensor as the roadway infrastructure is typically created for human visual perception. Computer vision comprises of semantic tasks such as object detection, semantic segmentation and geometric tasks like depth estimation, motion estimation and localization. In particular, for visual perception, Convolutional Neural Networks (CNNs) dominate, and are the standard models to perform bounding box object detection for detecting pedestrians, cyclists and vehicles. This application is quite mature and has already been deployed in previous generation vehicles for driver assistance and automated driving applications \cite{heimberger2017computer}. However neural networks are beginning to be applied in other areas of automated driving \cite{talpaert2019exploring}.

In the computer vision community, there is a strong trend of applying CNN for various computer vision tasks. Most of these efforts are focused on a single task, but due to the recent convergence of CNNs as the dominant models, there are few attempts to model multiple tasks in a unified way. 
For example, UberNet \cite{kokkinos2017ubernet} demonstrated joint learning of low, mid and high level computer vision tasks with high accuracy. We are motivated by this recent progress and propose a solution towards a unified perception model for all visual perception tasks in automated driving. As far as the authors are aware, there is no solution for this because of many practical challenges. The closest work done which is close to our objective is Magic Leap's unified model for perception tasks in augmented reality \cite{magicleap}.

The rest of the paper is structured as follows. Section \ref{vpad} reviews various visual perception tasks in automated driving and discusses CNN based solutions. Section \ref{universal} summarizes the arguments for a unified model and proposes a solution.  Section \ref{proposed} details the implementation of two-stream three-task networks and discusses experimental results of auxiliary learning and multi-stream networks. Finally, section \ref{conclusion} summarizes the paper and provides potential future directions.

\section{Automotive Visual Perception Tasks} \label{vpad}

\subsection{Object recognition}

\begin{figure*}
\centering
\includegraphics[width=0.7\textwidth]{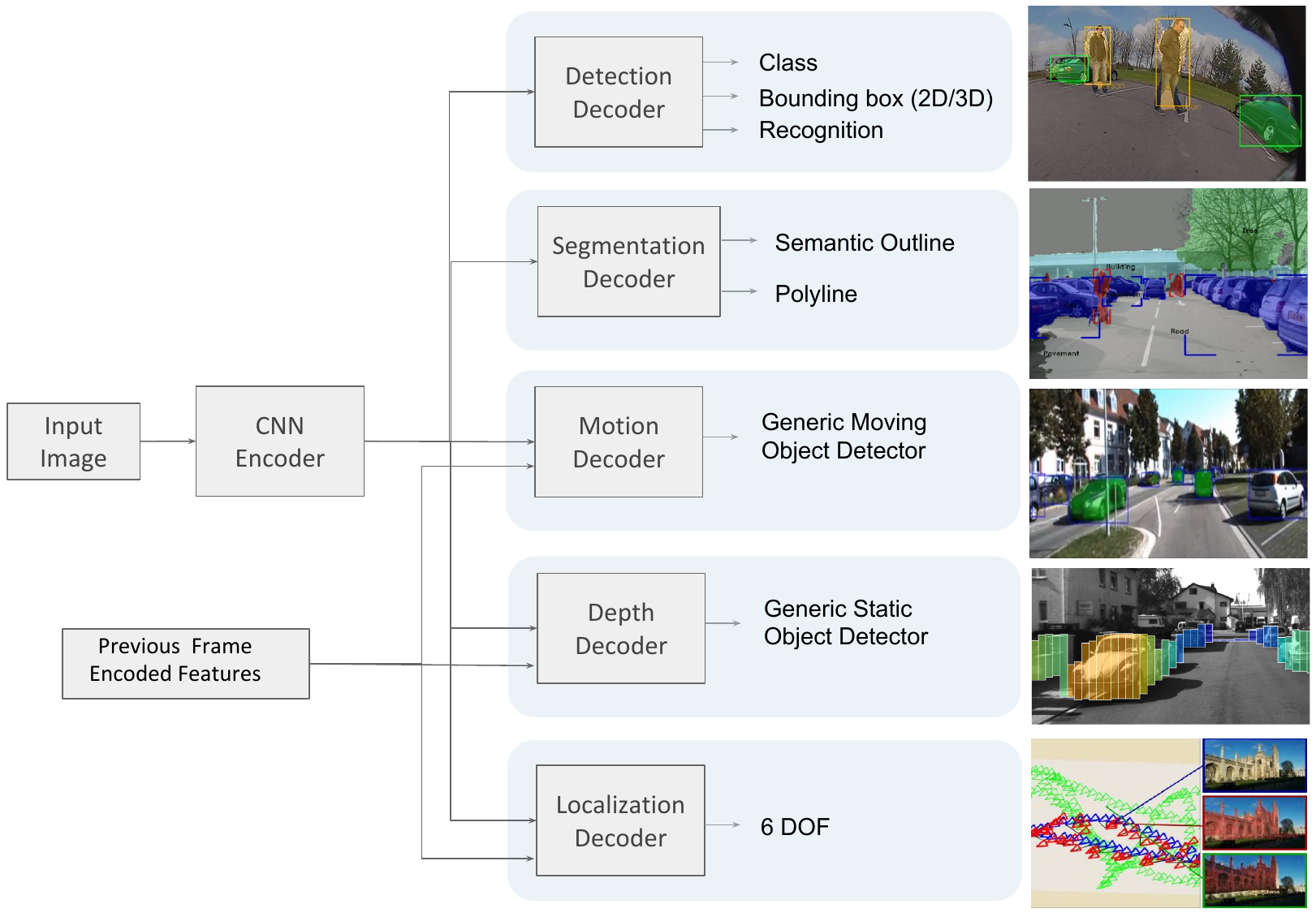}
\caption{NeurAll: Proposed unified CNN architecture for the important visual perception tasks in automated driving}
\label{fig:multi-task}
\end{figure*}



Object recognition is one of the first successful applications of deep learning for automated driving. The traditional approaches for bounding box object detection were replaced by CNN based approaches for pedestrians and vehicles. Although semantic segmentation can for certain applications supersede bounding box detection, there are many practical reasons for it to be used. 
Also, annotation for semantic segmentation is time consuming, typically taking up to an hour for annotating a single image. 
Additionally, the computational power of embedded system needed for deployment is still a bottleneck, as network sizes are larger for segmentation compared to bounding box, for the same performance, as the decoder needs to decode to full resolution. Thus we focus our development on a limited number of important classes in each of the categories. 
The first important group of classes is roadway objects being the static infrastructure of road, lane and curb. They cannot be represented by a bounding box and thus require a segmentation task. The second group of classes is the dynamic objects of vehicles, pedestrians and cyclists. 


Siam {\em et al.} \cite{siam2017deep} provided a detailed overview of semantic segmentation for automated driving applications. The majority of semantic segmentation algorithms follow an encoder/decoder-like architecture, where the encoder is a trimmed version of pre-trained classification network and decoder consists of series of deconvolutions \cite{noh2015learning} or upscaling layers \cite{7803544}. Skip connections between encoder and decoder layers are also well studied \cite{unet}. This encoder/decoder-like architecture encourages other tasks to be modeled in the same manner. The CNN based bounding box regression algorithms can be broadly categorized into two groups, single stage and two stage approaches. Single stage approaches, regress for box co-ordinates and class category in one shot. YOLO \cite{YOLOV3} and SSD \cite{SSD} are pioneering works in single stage methods. On the other hand two stage networks involve explicit loss functions for class agnostic region proposals followed by accurate box co-ordinates regression. The R-CNN family of algorithms \cite{ren2015faster} fall into this category. 



\subsection{Generic object detection}
It is not possible to list the entire set of objects in an automotive scenario and build a training dataset for appearance based classification.
Thus, there will always be objects that are not trained for, or that do not form part of the training dataset (e.g. kangaroos, moose, or obscure construction vehicles with distinctly unfamiliar visual appearances).
In some cases, trucks or buses with ads can confuse the object detection model. Thus appearance agnostic object detection is critical for automated driving systems, and alternate geometric cues of motion and depth are required. Even in case of standard objects like pedestrians and vehicles, such cues will aid the robustness of detection. 
In classical computer vision, motion is computed using optical flow and depth is computed using structure from motion. Then post processing algorithms, such as clustering, are performed to extract generic static and moving objects. CNNs have gradually shown high accuracy for estimation of dense optical flow \cite{ilg2017flownet} and of monocular depth \cite{ummenhofer2017demon}. To estimate flow and depth, it is necessary to exploit motion parallax using two consecutive frames. CNNs can also be used to extract moving and static objects directly instead of the intermediate pixel-level flow or depth estimation which is more complex. Moving Object Detection Network (MODNet) \cite{siam2017modnet} poses moving object detection as a binary segmentation problem and directly estimated. StixelNet \cite{garnett2017real} poses generic static obstacles represented as stixels and learnt directly using a CNN. MODNet and StixelNet demonstrated shared encoder for the standard object detection. 

\subsection{Odometry, Localization and Mapping}
In a typical automotive setup, ego-vehicle is constantly in motion and it is important to measure the relative pose change through Visual Odometry (VO). 
In general, there are two deep learning based approaches. The simplest one is to take a pair of images in a sequence and estimate VO using CNN. The second approach is to leverage the smoothness of the odometry by making use of a higher order temporal mode. Qiao {\em et al.} \cite{qiao2017learning} use a simple 2 frame estimation of VO by combining encoders of two consecutive frames and decoding them to estimate 6 DOF. Clark {\em et al.} \cite{clark2017vinet} jointly pose visual and inertial data VO as a sequence to sequence learning problem. Weber {\em et al.} \cite{weberlearning} explicitly model learning of temporal features. Wang {\em et al.} \cite{wang2017deepvo} use a combination of CNN and RNN to estimate VO. 

Visual SLAM is a key technology for automated driving and many fully automated driving prototypes (e.g., Google Waymo) are currently based on localization to pre-mapped regions. Visual SLAM is a classical computer vision and robotics discipline comprising of building blocks including feature tracking, mapping, global optimization and re-localization. CNN based methods are gradually replacing some of these components. Milz {\em et al.} \cite{milz2018visual} provide a detailed overview of current applications of and future directions for CNN based Visual SLAM for automated driving. In addition, re-localization to a high definition dense map provides a strong prior for static object detection and can estimate moving objects by background subtraction.

\subsection{Other tasks}

Automated driving has a variety of scenarios like highway driving, parking, urban driving, etc. These scenarios are typically classified using a CNN using a scene classification task. Other visual perception tasks include soiling detection, adverse weather detection, online calibration, etc. 
Soiling detection refers to detection of external contamination on automated driving cameras while adverse weather detection classifies various weather conditions like rainy, foggy, snowy, etc. Computer vision algorithms are sensitive to camera calibration errors, thus online camera calibration is needed to account for the mechanical wear and tear of the car body. In multicamera systems, it is even more important to accurately estimate calibration in order to provide a surround view output. There is very little literature on using deep learning for the tasks discussed in this section, with some exceptions. An example is the work of Itu {\em et al.} \cite{itu2017automatic} which uses CNN to estimate vanishing point for obtaining extrinsic calibration, and Sakaridis {\em et al.} \cite{Sakaridis_2018_ECCV} on foggy scene understanding. However, in principle, these tasks should be feasible within the unified NeurAll framework.

\section{Unified visual perception model} \label{universal}

\subsection{Motivation}

\subsubsection{Universality of CNN features}
As discussed in section \ref{vpad}, CNN approach has been successfully applied to various tasks due to the universality of features in many different aspects.
The main advantage of the universality of these features is that they are transferable to other tasks when previously learned for a different task, a widespread practice known as Transfer Learning. For example, weights obtained from a pre-trained network on the Imagenet \cite{deng2009imagenet} classification task are usually used as an initialization step for fine-tuning more complex vision tasks.
Transfer learning allows knowledge transfer from one task to another, mostly when the data are lacking for the targeted task. Indeed, the low level features are mostly task agnostic so that it can  be reused by other vision tasks. 
But how far can this practice be applied? To better understand this concept of transfer learning, Zamir {\em et al.} \cite{zamir2018taskonomy} characterized the relationship between several visual tasks and found an order of dependency for task transfer learning. Other  papers \cite{bilen2017universal,rebuffi2018efficient} demonstrated that universality not only applies to different domains but also to different modalities. Kaiser {\em et al.} \cite{kaiser2017one} jointly learned tasks in different modalities (speech recognition, image classification and text translation) using a single model. 
Therefore, feature sharing is possible for very different tasks across a shared network. 

Learning multiple tasks simultaneously using a shared representation is known as multi-task learning. Learning a universal representation to solve multiple tasks is a key step in the development of efficient algorithms in terms of performance, generalization and computational complexity instead of having several separate networks for different tasks. Recently, several works \cite{Teichmann2018MultiNetRJ,kokkinos2017ubernet,neven2017fast} proposed a joint multi-task network to solve several tasks simultaneously. However, universality of CNN features is only possible up to a certain extent. Indeed, one limitation of the CNN is their ease to specialize to a domain or task, preventing their generalization to other domains or tasks. In order to overcome this limitation, Bilen {\em et al.} \cite{bilen2017universal} normalized the information in the network using domain-specific scaling factors or normalization layers.
Rebuffi {\em et al.} \cite{rebuffi2018efficient} built a universal parametric families of networks for efficient sharing of parameters across domains.
Tamaazousti {\em et al.} \cite{tamaazousti2018universality} proposed universalizing methods to force a network to learn a representation capable of handling various tasks. These progresses suggest CNN features offer a strong possibility to represent multiple tasks through a unified model. 
\subsubsection{Adaptation of unified model}
Universal feature representation can fail for several reasons. For example, the representation capacity of the model for tasks with different complexities can be insufficient or oversized. In some cases, representations that are too generic might prevent the specialization to a certain task or domain. The complexity of automotive tasks can have a large variability in practise and thus it is necessary to have adaptation mechanisms to obtain optimal feature representation. Designing a balanced dataset satisfying complexities of different tasks is challenging. 
In general, complex tasks require bigger models while simple tasks need smaller models. An undersized representation for a complex task can be compensated by augmenting specialized feature maps just for the more complex tasks. In this case, these augmented features are used only for the complex tasks.
On the other hand, an oversized representation for a simpler task can be avoided by pruning to simplify the model. Pruning methods compress the model by removing redundant filters and keeping only the most relevant ones \cite{liu2017learning,huang2018condensenet}. Here we propose to perform task specific feature pruning for simpler tasks.
Thus a shared model constructed with these adaptations can then be effectively utilized by tasks with variable complexities.
 

\subsubsection{Hardware trend}
Computational power in embedded systems for deploying automated driving solutions is rapidly growing. In particular, there are specialized accelerators available for CNNs. Convolution is the main compute intensive operation in a CNN, offering heavy parallelism that is suitable for specialized hardware. Majority of the hardware vendors for automated driving have custom hardware accelerators for CNNs. For example, Nvidia Xavier \cite{nvidiaXavier} provides $\sim$30 TOPS (Tera Operations per second) for CNNs. Relatively, compute power available for general purpose processing is much lower and the trend shows that it will reduce even further. Thus even if there is a classical algorithm which is more optimal for tasks like calibration, a CNN based approach will be a more efficient mapping to these specialized hardwares.

\subsection{Pros and cons of unified model}

\subsubsection{Pros}
The main advantage of unified model is improving computational efficiency. Say there are two problems with two equivalent independent networks utilizing 50\% of available processing power. A unified model with 30\% sharing across the two networks can offer 15\% of additional resources to each network for computing a slightly larger problem. This allows unified models to offer scalability for adding new tasks at a minimal computation complexity. On other hand, these models reduce development effort and training time as shared layers minimize the need of learning multiple set of parameters in different models. Unified models learn features jointly for all tasks which makes them robust to over-fitting by acting as a regularizer, as demonstrated in various multi-task networks \cite{Teichmann2018MultiNetRJ,kokkinos2017ubernet,neven2017fast}.

\subsubsection{Cons}
In case of separate models, the algorithms are completely independent. This could make the dataset design, architecture design, tuning, hard negative mining, etc. simpler and easier to manage. Debugging a unified model can be quite challenging. These models are often less fault tolerant since features are shared for all tasks leading to a single point of failure. Such failures in learning features for a particular scenario might negatively impact other tasks. This is often called as negative transfer in multi-task learning. Another practical disadvantage is that unified models assume a fixed input format for all tasks and some tasks might need a different setting like camera field-of-view, color format or pixel resolution.


\subsection{Proposed baseline of unified model}


Figure \ref{fig:multi-task} illustrates a simple unified model architecture for five main tasks in automated driving. Throughout this paper, we refer to the shared layers in the unified model as CNN encoder and task dependent layers as decoders. It is straight forward to add other tasks like calibration or depth estimation via additional task specific decoders. This architecture has a shared CNN encoder and multiple parallel task dependent decoders.

In this unified model, object detection decoders like YOLO \cite{YOLOV3}, SSD \cite{SSD} can be used to predict bounding boxes and categories of objects while segmentation decoder like FCN8 \cite{long2015fully} can be used to perform pixel wise semantic segmentation. Motion and depth decoders can be constructed to learn simpler representations for generic object detection. A motion decoder can be used to perform binary segmentation of moving objects and depth decoder to estimate variable height stixels of static objects. Finally, a localization decoder can be used to predict pose of camera similar to PoseNet \cite{kendall2015posenet}. Multinet by Teichmann {\em et al.} \cite{Teichmann2018MultiNetRJ} and fast scene understanding by Neven {\em et al.} \cite{neven2017fast} share similar design but for fewer tasks without multi-stream design explained later. 

In order to facilitate joint training of unified model, tasks of varying complexities and  datasets have to be balanced. It is challenging to collect annotated data for all tasks. To alleviate this problem in a heterogeneous dataset, UberNet \cite{kokkinos2017ubernet} proposes an asynchronous variant of backpropagation where training data are sequentially read and task specific parameters are updated only after accumulating sufficient training examples for this particular task.

\subsection{Improvements over the baseline model} 

\subsubsection{Task loss weighting} Another key challenge in unified model training is to perform a well-balanced optimization of the $\mathnormal{N}$ tasks, so that the features of each task are learned equally in a shared representation. Most of the papers express multi-task loss as a weighted sum of the individual task losses. 
 Incorrect setting of weight parameters for task losses could lead to negative transfer in multi-task learning. Kendall {\em et al.} \cite{kendall2018multi} learned optimal weights via uncertainty of tasks. Chen {\em et al.} \cite{Chen2018GradNormGN} achieved adaptive weighting by penalizing the network on higher magnitudes of back-propagated gradients. Guo {\em et al.} \cite{guo2018dynamic} updated tasks weights by observing difficulty of the tasks during training. 
 Thus, it is advised to use appropriate tuning of weight parameters for success in multi-task learning. 
\begin{figure*}[!t]
\centering
\includegraphics[width=0.8\textwidth]{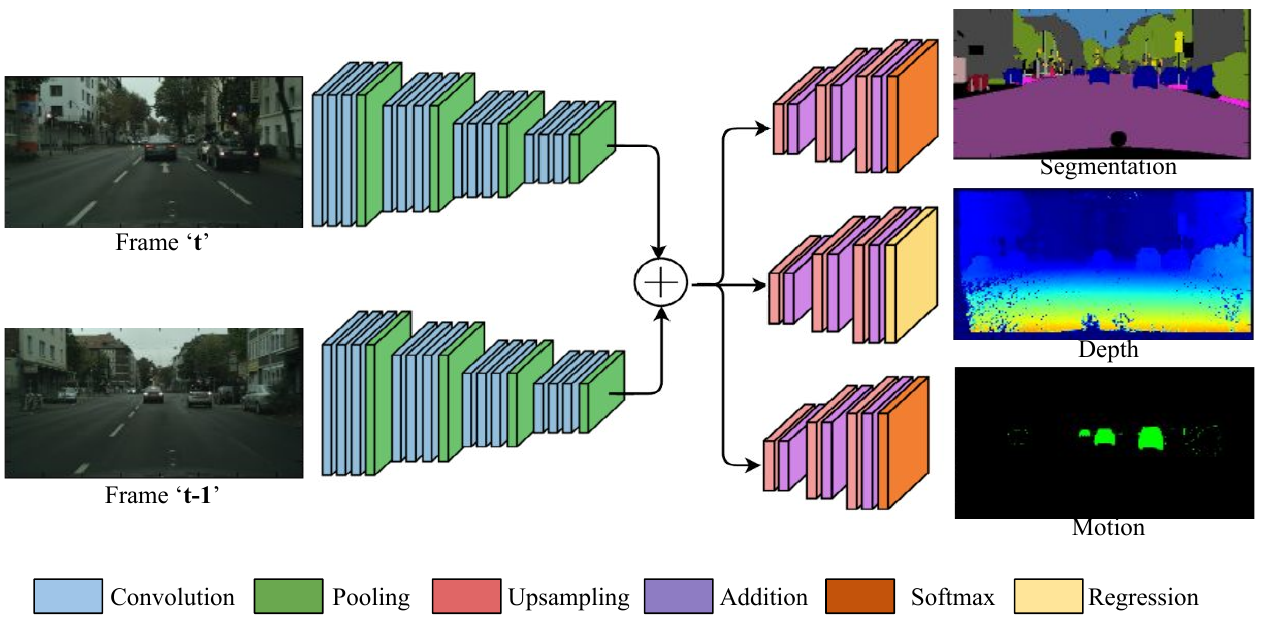}
\caption{Two-stream three-task visual perception network architecture performing segmentation, depth estimation and moving object detection}

\label{fig:two-task}
\end{figure*}

\subsubsection{Multi-stream learning}
\label{multi-stream}
Prior works \cite{Simonyan2014TwoStreamCN,Vertens2017SMSnetSM} 
show that motion and depth cues can be obtained from two-stream architectures. Adding a secondary stream to process previous frames in a sequence helps identifying motion cues similarly processing left/right images in a stereo pair can provide depth information. In this fashion, unified model can be extended by adding multiple streams to process inputs from different cameras of a 360$\degree$ camera setup. The CNN encoder weights can be reused between multiple streams allowing better computational efficiency.

\subsubsection{Auxiliary learning}
Auxiliary learning is another technique to improve multi-task learning performance. Learning an auxiliary task that is nonoperational during inference with a prime motivation to enhance main task performance is referred to as auxiliary learning. There are many tasks such as the prediction of steering angle which are free without any annotation cost and can be used as auxiliary tasks. This auxiliary task is usually selected to have much larger annotated data so that it acts a regularizer for main task. Liebel {\em et al.} \cite{liebel2018auxiliary} performed semantic segmentation using auxiliary tasks like time, weather, etc. It is often believed that auxiliary tasks can be used to focus attention on a specific parts of the input. Predictions of road characteristics like markings as an auxiliary task \cite{caruana1997multitask} to improve main task for steering prediction is one instance of such behaviour. 

\subsubsection{Synergized decoders}
In the baseline model, the decoders for different tasks are independent. However, there could be possible synergies across the tasks. Human visual system combines motion, depth and appearance to re-focus and detect objects adaptively. Inspired by this, we propose to have cross decoder connections which are learnt from data. For example, depth could help in obtaining better boundaries for recognition of objects and knowing that certain pixels are road can provide priors for depth estimation.

\section{Implementation and results} \label{proposed}

There are many challenges in getting the full NeurAll model implemented. Firstly, there are no datasets providing simultaneous annotation for all the five tasks proposed in Fig. \ref{fig:multi-task}. Secondly, there are practical limitations with GPU memory for training a complex model. Thus we implement a three-task model, illustrated in Fig. \ref{fig:two-task}, as a first step towards NeurAll. Due to limited space, the details and experiments are kept brief and will be expanded up in a separate paper. In future work, we aim to make a public dataset providing annotation for ten visual perception tasks so that a larger unified model can be evaluated.

\subsection{Multi-stream learning}
As discussed earlier
, multi-stream can  be used for depth regression and optical flow estimation. In order to incorporate temporal cues into the unified model, we constructed two multi-stream models that process consecutive frames sampled from a video sequence. Using a shared encoder in these models, we minimize the computational complexity. The encoded features from consecutive frames can be simply concatenated or processed by a recurrent unit and then passed to a segmentation decoder. This allows video segmentation instead of naive frame by frame processing. To demonstrate benefits of multi-stream models, we constructed a baseline single task segmentation model (SegNet) using ResNet-50 \cite{7780459} encoder (with pre-trained weights) and FCN8 \cite{long2015fully} decoder (randomly initialized weights) and trained this model for segmentation task. Our multi-stream models MSNet$_2$ and RNNet$_2$ are built on baseline SegNet with an additional input stream. We share the encoder between two streams in both models. The main difference between these models is that encoded features from consecutive frames are simply concatenated in MSNet$_2$ while they are processed by a Long Short Term Memory Unit (LSTM) in RNNet$_2$. All weight parameters of these models using ResNet-50 \cite{7780459} encoder and FCN8 \cite{long2015fully} decoder are learned in a similar fashion to the two-task model in previous section.
\begin{table*}[]
\centering
\caption{multi-stream Semantic Segmentation Results on KITTI and SYNTHIA Video Sequences. SegNet: Single frame, MSNet$_{2}$: multi-stream with 2 frames input, RNNet$_{2}$: Recurrent unit with 2 frames input. Shared encoder is used for multi-stream inputs. }
\label{table:multi-stream}
\resizebox{0.925\textwidth}{!}{%
\begin{tabular}{lccccccccccc}
\hline
\multicolumn{12}{c}{\textbf{KITTI}} \\ \hline
\textbf{Model} & \textbf{Sky} & \textbf{Building} & \textbf{Road} & \textbf{Sidewalk} & \textbf{Fence} & \textbf{Vegetation} & \textbf{Pole} & \textbf{Car} & \textbf{Lane} & \textbf{IoU} & \textbf{Params} \\ \hline
SegNet & 46.79 & 87.32 & 89.05 & 60.69 & 22.96 & 85.99 & - & 74.04 & - & 74.52 & 23,668,680 \\
MSNet$_{2}$ & 47.89 & 91.08 & \textbf{97.58} & 88.02 & \textbf{62.60} & \textbf{92.01} & - & \textbf{90.26 }& - & \textbf{85.31} &23,715,272\\
RNNet$_{2}$ & \textbf{50.20} & \textbf{93.74} & 94.90 & \textbf{88.17} & 59.73 & 87.73 & - & 87.66 & - & 84.19 & 31,847,828 \\ 
\hline
\hline
\multicolumn{12}{c}{\textbf{SYNTHIA}} \\ \hline
\textbf{Model} & \textbf{Sky} & \textbf{Building} & \textbf{Road} & \textbf{Sidewalk} & \textbf{Fence} & \textbf{Vegetation} & \textbf{Pole} & \textbf{Car} & \textbf{Lane} & \textbf{IoU} & \textbf{Params} \\ \hline
SegNet & \textbf{95.41} & 58.18 & 93.46 & 09.82 & 76.04 & 80.95 & 08.79 & 85.73 & 90.28 & 89.70 & 23,692,820 \\
MSNet$_{2}$  & 92.43 & 66.12 & 94.21 & 28.17 & \textbf{84.87} & 78.01 & 46.78 & 82.14 & \textbf{95.26} & 93.44 & 23,739,412 \\
RNNet$_{2}$ & 94.17 & \textbf{71.48}  & \textbf{95.16} & \textbf{30.94} & 82.26 & \textbf{82.76} & \textbf{47.28} & \textbf{87.03} & 93.97 & \textbf{94.17} & 31,861968\\ \hline
\end{tabular}%
}
\end{table*}

\begin{table*}[]
\centering
\caption{Comparison Study : Single task vs Auxiliary learning. AuxNet$_{400}$ and AuxNet$_{1000}$ weighs segmentation loss 400 and 1000 times compared to depth loss. AuxNet$_{\rm TWB}$ is constructed by expressing total loss as product of task losses. }
\label{table:auxnet}
\resizebox{0.9\textwidth}{!}{%
\begin{tabular}{lcccccccccc}
\hline
\multicolumn{11}{c}{\textbf{KITTI}} \\ \hline
\textbf{Model} & \textbf{Sky} & \textbf{Building} & \textbf{Road} & \textbf{Sidewalk} & \textbf{Fence} & \textbf{Vegetation} & \textbf{Pole} & \textbf{Car} & \textbf{Lane} & \textbf{IoU} \\ \hline
SegNet & 46.79 & 87.32 & 89.05 & 60.69 & \textbf{22.96} & 85.99 & - & 74.04 & - & 74.52 \\
AuxNet$_{400}$ & 49.11 & 88.55 &\textbf{ 93.17 }& 69.65 & 22.93 & \textbf{87.12} & - & 74.63 & - & 78.32 \\
AuxNet$_{1000}$ & 49.17 & 89.81 & 90.77 & 64.16 & 14.77 & 86.52 & - & 71.40 & - & 76.58  \\ 
AuxNet$_{\rm TWB}$ & \textbf{49.73} & \textbf{91.10} & 92.30 & \textbf{70.55} & 18.64 & 86.01 & - &\textbf{ 77.32} & - & \textbf{78.64}
\\ 
\hline
\hline
\multicolumn{11}{c}{\textbf{SYNTHIA}} \\ \hline
\textbf{Model} & \textbf{Sky} & \textbf{Building} & \textbf{Road} & \textbf{Sidewalk} & \textbf{Fence} & \textbf{Vegetation} & \textbf{Pole} & \textbf{Car} & \textbf{Lane} & \textbf{IoU}  \\ \hline
SegNet & \textbf{95.41} & 58.18 & 93.46 & 09.82 & 76.04 & 80.95 & 08.79 & 85.73 & 90.28 & 89.70 \\
AuxNet$_{400}$ & 95.12 & \textbf{69.82} & 92.95 & 21.38 & 77.61 & 84.23 & 51.31 & \textbf{90.42} & 91.20 & 91.44 \\
AuxNet$_{1000}$ & \textbf{95.41} & 59.57 & \textbf{96.83} & 28.65 & \textbf{81.23} & 82.48 & \textbf{56.43} & 88.93 & 94.19 & \textbf{92.60}\\ 
AuxNet$_{\rm TWB}$ & 94.88 & 66.41 & 94.81 & \textbf{31.24} & 77.01 & \textbf{86.04} & 21.83 & 90.16 & \textbf{94.47 }& 91.67 
\\\hline
\end{tabular}%
}
\end{table*}
Table \ref{table:multi-stream} compares performance of multi-stream models used for video segmentation with single stream model. It is observed that the multi-stream models outperform single stream semantic segmentation at very minimal increase in computational complexity. Inducing valuable temporal cues by simple concatenation of encoded features from consecutive frames in MSNet$_2$ offer a replacement to compute intensive LSTM unit used in RNNet$_2$. We could not perform experiments on unified model with multi-stream inputs due to unavailability of dataset with both segmentation and detection labels for a video sequence. However, multi-stream models demonstrate their potential improvements to a unified model by achieving 11\% and 4\% gains in terms of IoU over a single task model for segmentation task on KITTI  \cite{geiger2013vision} and SYNTHIA \cite{ros2016synthia} validation sets.

\subsection{Auxiliary learning}
We conducted experiments to evaluate improvements offered by auxiliary learning to a unified model by choosing depth regression as an auxiliary task for main task semantic segmentation. 
We use SegNet as baseline to study the improvements of auxiliary learning. We built our auxiliary learning network (Auxnet) by adding a depth regression decoder to the baseline SegNet model keeping the ResNet-50 \cite{7780459} encoder common for segmentation and depth regression decoders. Depth regression decoder was constructed similar to FCN8 \cite{long2015fully} decoder except the final layer was replaced with regression units instead of softmax to estimate depth. Similar to previous experiments, we used pre-trained weights for ResNet-50 \cite{7780459} encoder and randomly initialized the weights for segmentation and depth decoders. Results of comparison between baseline semantic segmentation (SegNet) and auxiliary learning model (AuxNet) are listed in Table \ref{table:auxnet}. We experimented with different hand-weighted and adaptive weighted task loss networks to understand the behaviour of auxiliary learning. We expressed total loss for adaptive weighted task loss as geometric mean of individual task losses. This adds a constraint for join minimization.  Auxiliary network achieved 4\% and 3\% IoU improvement on KITTI \cite{geiger2013vision} and SYNTHIA \cite{ros2016synthia} validation sets.



\subsection{Two-stream three-task unified model vs. independent single task models}


We conducted a simple experiment to show the benefits of using a multi-task unified model vs. several single task models.
In particular, we focused on three tasks namely semantic segmentation, depth estimation and motion detection. 
On one hand, we built a three task unified model consisting of a dual identical encoders for the current and previous frames, and three individual decoders, performing segmentation, motion detection and depth respectively. For comparison, we built three separate single task decoders, with single encoder for each.
We used a ResNet-50 encoder \cite{7780459} pretrained on ImageNet and fine-tuned on the dataset. FCN8 \cite{long2015fully} is used as the decoder in all three cases, using final softmax layers for segmentation and motion detection, and a regression layer for depth (Figure \ref{fig:two-task}).


For scalarization of the three loss functions, we use the standard sum of the losses as a baseline and improve it by using product of losses inspired by its use in game theory \cite{miettinen2002scalarizing}. For the three task problem with task losses $`\mathcal{L}_{1}$',`$\mathcal{L}_{2}$' `$\mathcal{L}_{3}$', we express total loss as:
\begin{equation}
\label{eq:ALS-3task}
    \mathcal{L}_{Total} =  \sqrt[3]{\mathcal{L}_{1}\mathcal{L}_{2}\mathcal{L}_{3}}
\end{equation}

We trained and evaluated the 
2-stream 3-task unified model and 3 single task models on different driving datasets: the publicly available datasets KITTI \cite{geiger2013vision} and Cityscapes \cite{cordts2016cityscapes}.
Categorical cross-entropy was used as loss metric for segmentation and motion detection whil huber loss is used as loss metric for depth estimation. Mean class IoU (Intersection over Union) and per-class IoU were used as accuracy metrics for semantic segmentation, motion detection. Regression accuracy was used as accuracy metric for depth estimation. 


\begin{figure*}[!t]
\centering
\includegraphics[width=.875\textwidth,height=6.75cm]{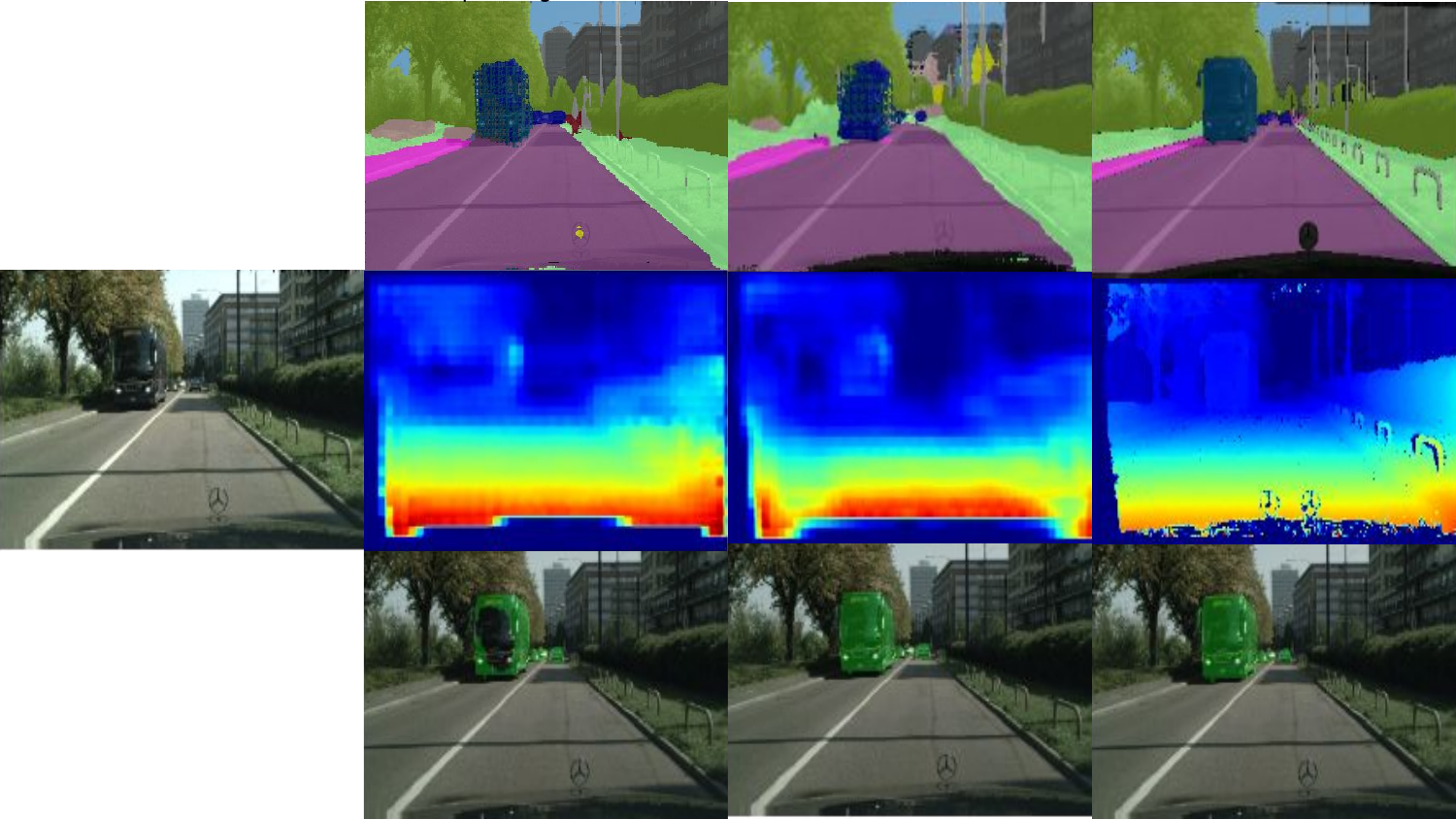}
\caption{Qualitative results of two-stream three-task model performing segmentation, depth estimation and motion segmentation. Left to Right:  Input Image, Single Task Network outputs, Multitask Output, Ground Truth. }
\label{fig:two-task-output}
\end{figure*}

\begin{table*}[ht]
\normalfont
\centering
\caption{Comparison of two-stream three-task model performing segmentation, depth estimation and motion segmentation on KITTI and Cityscapes datasets. Results of One-task and two-task networks are provided for comparison.}
\resizebox{0.75\textwidth}{!}{%
\begin{tabular}{l|ccc|ccc}
\hline
\multicolumn{1}{c|}{\multirow{2}{*}{Method}} & \multicolumn{3}{c|}{KITTI} & \multicolumn{3}{c}{Cityscapes} \\
\multicolumn{1}{c|}{} & Segmentation & Depth & Motion & Segmentation & Depth & Motion \\ 
\hline \hline
\multicolumn{7}{c}{\textbf{1-Task Segmentation, Depth or Motion}} \\\hline 
1-Task & 81.74\% & 75.91\% & 98.49\% & 78.95\% & 60.13\% & 98.72\% \\
\hline \hline
\multicolumn{7}{c}{\textbf{2-Task Segmentation and Depth}} \\\hline
Sum of Losses & 74.30\% & 74.47\% & - & 73.76\% & 59.38\% & - \\
Product of Losses & 81.50\% & 74.92\% & - & 79.14\% & 60.15\% & - \\
\hline \hline
\multicolumn{7}{c}{\textbf{2-Task Segmentation and Motion}} \\\hline
Sum of Losses & 80.14\% & - & 97.88\% & 78.46\% & - & 98.25\% \\
Product of Losses & 81.52\% & - & 97.93\% & 77.63\% & - & 98.83\% \\
\hline \hline
\multicolumn{7}{c}{\textbf{3-Task Segmentation, Depth and Motion}} \\\hline
Sum of Losses & 77.14\% & 76.15\% & 97.83\% & 72.71\% & 60.97\% & 98.20\% \\
Product of Losses &82.20\% & 76.54\% & 97.92\% & 77.38\% & 61.56\% & 98.72\% \\
\hline
\end{tabular}%
}
\vspace{-0.5cm}
\label{tab:main-results}
\end{table*}

Figure \ref{fig:two-task-output} illustrates qualitative output of the two-stream 3-task network and Table \ref{tab:main-results} summarizes the obtained results.
We compare the results of 2-task models and 3-task models using product of losses against naive sum of losses (equal task weight method), and compare their performances with 1-task segmentation, depth and motion models. Our product of losses method shows significant improvements in performance over sum of losses method in both 2-task and 3-task models. This experiment shows the capacity of the unified model to learn multiple tasks with similar accuracies compared to single task models having significantly fewer parameters by choosing the appropriate task loss strategy. The segmentation, depth and motion single task models require 23.77M, 23.59M and 23.60M parameters respectively. In contrast, both two-task models required 23.86M parameters and the three task model required just 23.87M parameters. This shows that we have a drastic gain in terms of memory and computational efficiency. The results here are preliminary and will be expanded upon in another paper.



\section{Conclusion} \label{conclusion}
CNNs have become the standard model for visual semantic tasks like object detection and semantic segmentation. CNNs are rapidly progressing to achieve state-of-the-art results for geometric tasks like depth estimation and visual SLAM. This brings an opportunity for CNN to become a unifying model for all visual perception tasks in automated driving. In this paper, we argue for moving towards a unified model and use current literature to propose how to achieve it. We also discuss the pros and cons of having a unified model. Finally, we perform experiments on a simpler scenario with two tasks and demonstrate results to support our argument. We demonstrated that usage of appropriate task weighted losses enhance the performance of multi-task learning. Using a shared encoder for multiple streams of inputs increases performance keeping the computational complexity lower. We hope that this work encourages further research in this area to enable a competitive solution for a unified model.



\bibliographystyle{unsrt}
\bibliographystyle{ieee}
\bibliography{egbib}

\end{document}